\pdfoutput=1
\documentclass[11pt]{article}

\usepackage[]{emnlp2021}

\usepackage{times}
\usepackage{latexsym}
\usepackage{bm}

\usepackage[T1]{fontenc}
\usepackage[utf8]{inputenc}
\usepackage[hang,flushmargin]{footmisc}

\usepackage{url}
\usepackage{amsmath}
\usepackage{tabularx}
\usepackage{graphicx}

\usepackage{microtype}
\usepackage{natbib}

\title{Aligning Cross-lingual Sentence Representations with \\ Dual Momentum Contrast}

\author{Liang Wang \and Wei Zhao \and Jingming Liu \\
        Yuanfudao AI Lab, Beijing, China \\
        \texttt{ \{wangliang01,zhaowei01,liujm\}@yuanfudao.com } }

\begin{document}
\maketitle
\begin{abstract}
In this paper,
we propose to align sentence representations from different languages
into a unified embedding space,
where semantic similarities (both cross-lingual and monolingual)
can be computed with a simple dot product.
Pre-trained language models are fine-tuned with the translation ranking task.
Existing work ~\citep{feng2020language} uses sentences within the same batch as negatives,
which can suffer from the issue of easy negatives.
We adapt MoCo ~\citep{he2020momentum} to further improve the quality of alignment.
As the experimental results show,
the sentence representations produced by our model
achieve the new state-of-the-art on several tasks,
including Tatoeba en-zh similarity search ~\citep{Artetxe2019MassivelyMS},
BUCC en-zh bitext mining,
and semantic textual similarity on 7 datasets.
\end{abstract}

\section{Introduction}

Pre-trained language models like BERT ~\citep{devlin-etal-2019-bert}
and GPT ~\citep{Radford2018ImprovingLU}
have achieved phenomenal successes on a wide range of NLP tasks.
However,
sentence representations for different languages are not very well aligned,
even for pre-trained multilingual models
such as mBERT ~\citep{Pires2019HowMI, Wang2020CrosslingualAV}.
This issue is more prominent for language pairs from different families
(e.g., English versus Chinese).
Also,
previous work ~\citep{li2020sentence} has shown that the out-of-box BERT embeddings
perform poorly on monolingual semantic textual similarity (STS) tasks.

There are two general goals for sentence representation learning:
cross-lingual representations should be aligned,
which is a crucial step for tasks
like bitext mining ~\citep{artetxe2019margin},
unsupervised machine translation ~\citep{Lample2018PhraseBasedN},
and zero-shot cross-lingual transfer ~\citep{Hu2020XTREMEAM} etc.
Another goal is to induce a metric space,
where semantic similarities can be computed with simple functions
(e.g., dot product on $L_2$-normalized representations).

Translation ranking ~\citep{feng2020language,Yang2020MultilingualUS}
can serve as a surrogate task to align sentence representations.
Intuitively speaking,
parallel sentences should have similar representations and are therefore ranked higher,
while non-parallel sentences should have dissimilar representations.
Models are typically trained with in-batch negatives,
which need a large batch size to alleviate the \emph{easy negatives} issue ~\citep{chen2020simple}.
~\citet{feng2020language} use \emph{cross-accelerator negative sampling}
to enlarge the batch size to $2048$ with $32$ TPU cores.
Such a solution is hardware-intensive and still struggles to scale.

Momentum Contrast (MoCo) ~\citep{he2020momentum} decouples the batch size
and the number of negatives by maintaining a large memory queue and a momentum encoder.
MoCo requires that queries and keys lie in a shared input space.
In self-supervised vision representation learning,
both queries and keys are transformed image patches.
However,
for translation ranking task,
the queries and keys come from different input spaces.
In this paper,
we present \emph{dual momentum contrast} to solve this issue.
Dual momentum contrast maintains two memory queues
and two momentum encoders for each language.
It combines two contrastive losses by performing bidirectional matching.

We conduct experiments on the English-Chinese language pair.
Language models that are separately pre-trained for English and Chinese
are fine-tuned using translation ranking task with dual momentum contrast.
To demonstrate the improved quality of the aligned sentence representations,
we report state-of-the-art results
on both cross-lingual and monolingual evaluation datasets:
Tatoeba similarity search dataset (accuracy $95.9\% \rightarrow 97.4\%$),
BUCC 2018 bitext mining dataset (f1 score $92.27\% \rightarrow 93.66\%$),
and 7 English STS datasets (average Spearman's correlation $77.07\% \rightarrow 78.95\%$).
We also carry out several ablation studies to help understand
the learning dynamics of our proposed model.

\section{Method}

\begin{figure}[ht]
\begin{center}
 \includegraphics[width=0.99\linewidth]{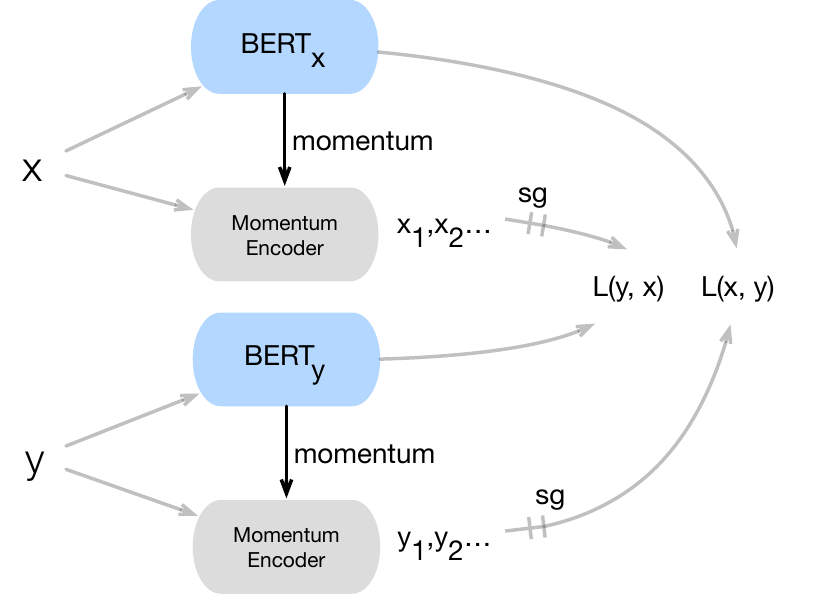}
 \caption{Illustration of dual momentum contrast.
 \emph{sg} denotes ``stop gradient''.
 \emph{x} and \emph{y} are sentences from two different languages.}
 \label{fig:model_arch}
\end{center}
\end{figure}

\noindent
\textbf{Dual Momentum Contrast }
is a variant of the MoCo proposed by ~\citet{he2020momentum}.
Our method fits into the bigger picture of contrastive learning
for self-supervised representation learning ~\citep{le2020contrastive}.
Given a collection of parallel sentences $\{x_i, y_i\}_{i=1}^n$,
as illustrated in Figure ~\ref{fig:model_arch},
we first encode each sentence using language-specific BERT models (base encoder),
then apply mean pooling on the last-layer outputs
and $L_2$ normalization to get the representation vector
$\mathbf{h}_{x_i}, \mathbf{h}_{y_i} \in R^{768}$.

Each BERT encoder has a momentum encoder,
whose parameters $\bm{\theta}$ are updated
by exponential moving average of the base encoder as follows:
\begin{equation}
    \bm{\theta}_t \leftarrow m \bm{\theta}_{t-1} + (1-m) \bm{\theta}_{\text{base}}
\end{equation}
Where $t$ is the iteration step.
Two memory queues are maintained for each language
to store $K$ vectors encoded by the corresponding momentum encoder
from most recent batches.
The oldest vectors are replaced with the vectors from the current batch
upon each optimization step.
The momentum coefficient $m \in [0,1]$ is usually very close to $1$ (e.g., $0.999$)
to make sure the vectors in the memory queue are consistent across batches.
$K$ can be very large (\textgreater $10^5$) to provide
enough negative samples for learning robust representations.

To train the encoders,
we use the InfoNCE loss ~\citep{oord2018representation}:
\begin{equation} \label{equ:infonce}
    \text{L}(x, y) = -\log \frac{\text{exp}(\mathbf{h}_x\cdot \mathbf{h}_y/\tau)}{\sum_{i=0}^K{\text{exp}(\mathbf{h}_x\cdot \mathbf{h}_{y_i}/\tau)}}
\end{equation}
$\tau$ is a temperature hyperparameter.
Intuitively,
Equation ~\ref{equ:infonce} is a (K+1)-way softmax classification,
where the translation sentence $y=y_0$ is the positive,
and the negatives are those in the memory queue $\{y_i\}_{i=1}^K$.
Note that the gradients do not back-propagate
through momentum encoders nor the memory queues.

Symmetrically,
we can get $\text{L}(y, x)$.
The final loss function is the sum:
\begin{equation}
    \min\ \ \text{L}(x, y) + \text{L}(y, x)
\end{equation}

After the training is done,
we throw away the momentum encoders and the memory queues,
and only keep the base encoders to compute the sentence representations.
In the following,
our model is referred to as MoCo-BERT.
\newline

\noindent
\textbf{Application }
Given a sentence pair ($x_i$, $y_j$) from different languages,
we can compute cross-lingual semantic similarity by taking
dot product of $L_2$-normalized representations $\mathbf{h}_{x_i}\cdot \mathbf{h}_{y_j}$.
It is equivalent to cosine similarity,
and closely related to the Euclidean distance.

Our model can also be used to compute monolingual semantic similarity.
Given a sentence pair ($x_i$, $x_j$) from the same language,
assume $y_j$ is the translation of $x_j$,
if the model is well trained,
the representations of $x_j$ and $y_j$ should be close to each other:
$\mathbf{h}_{x_j} \approx \mathbf{h}_{y_j}$.
Therefore,
we have $\mathbf{h}_{x_i}\cdot \mathbf{h}_{x_j} \approx \mathbf{h}_{x_i}\cdot \mathbf{h}_{y_j}$,
the latter one is cross-lingual similarity
which is what our model is explicitly optimizing for.

\section{Experiments}

\subsection{Setup}

\noindent
\textbf{Data }
Our training data consists of English-Chinese corpora from
UNCorpus~\footnote{\url{https://conferences.unite.un.org/uncorpus}}, Tatoeba,
News Commentary~\footnote{\url{https://opus.nlpl.eu/}},
and corpora provided by CWMT 2018~\footnote{\url{http://www.cipsc.org.cn/cwmt/2018/}}.
All parallel sentences that appear in the evaluation datasets are excluded.
We sample 5M sentences to make the training cost manageable.

\noindent
\textbf{Hyperparameters }
The encoders are initialized with \emph{bert-base-uncased} (English) for fair comparison,
and \emph{RoBERTa-wwm-ext}~\footnote{\url{https://github.com/ymcui/Chinese-BERT-wwm}}(Chinese version).
Using better pre-trained language models is orthogonal to our contribution.
Following ~\citet{Reimers2019SentenceBERTSE},
sentence representation is computed by the mean pooling of the final layer's outputs.
Memory queue size is $409600$,
temperature $\tau$ is $0.04$,
and the momentum coefficient is $0.999$.
We use AdamW optimizer with maximum learning rate $4\times10^{-5}$ and cosine decay.
Models are trained with batch size $1024$
for $15$ epochs on $4$ V100 GPUs.
Please checkout the Appendix ~\ref{app:setup} for more details about data and hyperparameters.

\subsection{Cross-lingual Evaluation}

\begin{table}[ht]
\centering
\scalebox{0.9}{\begin{tabular}{l|c}
\hline
Model  & Accuracy \\ \hline
m$\text{BERT}_{\text{base}}$ ~\citep{Hu2020XTREMEAM} &    71.6\%   \\ \hline
LASER ~\citep{Artetxe2019MassivelyMS}  &    95.9\%  \\ \hline
VECO ~\citep{luo2020veco} &    82.7\%  \\ \hline
$\text{SBERT}_{\text{base}}$-p $^\dagger$      &   95.0\%   \\ \hline \hline
MoCo-$\text{BERT}_{\text{base}}$ (zh$\rightarrow$en) &  \textbf{97.4\%}  \\ \hline
MoCo-$\text{BERT}_{\text{base}}$ (en$\rightarrow$zh) &  96.6\%   \\ \hline
\end{tabular}}
\caption{Accuracy on the test set of Tatoeba en-zh language pair.
$\dagger$: ~\citet{reimers2020making}.}
\label{tab:tatoeba}
\end{table}

\begin{table}[ht]
\centering
\scalebox{0.9}{\begin{tabular}{l|c}
\hline
Model     & F1 \\ \hline
m$\text{BERT}_{\text{base}}$ ~\citep{Hu2020XTREMEAM}    &  50.0\%    \\ \hline
LASER ~\citep{Artetxe2019MassivelyMS}    &   92.27\%    \\ \hline
VECO ~\citep{luo2020veco}    &   78.5\%   \\ \hline
S$\text{BERT}_{\text{base}}$-p$^\dagger$ &  87.8\%    \\ \hline
LaBSE ~\citep{feng2020language}   &   89.0\%    \\ \hline \hline
MoCo-$\text{BERT}_{\text{base}}$   &  \textbf{93.66\%}    \\ \hline
\end{tabular}}
\caption{F1 score on the en-zh test set of BUCC 2018 dataset.
$\dagger$: ~\citet{reimers2020making}.}
\label{tab:bucc}
\end{table}

\noindent
\textbf{Tatoeba cross-lingual similarity search }
Introduced by ~\citet{Artetxe2019MassivelyMS},
Tatoeba corpus consists of $1000$ English-aligned sentence pairs.
We find the nearest neighbor for each sentence in the other language
using cosine similarity.
Results for both forward and backward directions are listed
in Table ~\ref{tab:tatoeba}.
MoCo-BERT achieves an accuracy of $97.4\%$.

\noindent
\textbf{BUCC 2018 bitext mining }
aims to identify parallel sentences from a collection of
sentences in two languages ~\citep{zweigenbaum2018overview}.
Following ~\citet{artetxe2019margin},
we adopt the margin-based scoring
by considering the average cosine similarity of $k$ nearest neighbors
($k=3$ in our experiments):
\begin{equation}
\begin{aligned}
    \text{sim}&(x, y)=\text{margin}(\text{cos}(x, y), \\
    & \sum_{z\in \text{NN}_k{(x)}} \frac{\text{cos}(x, z)}{2k} +  \sum_{z\in \text{NN}_k{(y)}} \frac{\text{cos}(y, z)}{2k})
\end{aligned}
\end{equation}

We use the distance margin function: $\text{margin}(a, b) = a - b$,
which performs slightly better than the ratio margin function ~\citep{artetxe2019margin}.
All sentence pairs with scores larger than threshold $\lambda$ are identified as parallel.
$\lambda$ is searched based on the validation set.
The F1 score of our system is $93.66\%$,
as shown in Table ~\ref{tab:bucc}.

\subsection{Monolingual STS Evaluation}

\begin{table*}[ht]
\centering
\scalebox{0.98}{\begin{tabular}{ccccccccc}
\hline
\multicolumn{1}{l|}{Model} & STS-12 & STS-13 & STS-14 & STS-15 & STS-16 & STS-B & \multicolumn{1}{c|}{SICK-R} & Avg \\ \hline
\multicolumn{9}{l}{\textit{w/o labeled NLI supervision}}         \\ \hline
\multicolumn{1}{l|}{Avg GloVe$^\dagger$}   &  55.14  & 70.66  &  59.73  & 68.25 &  63.66  &   58.02  & \multicolumn{1}{c|}{53.76} & 61.32  \\ \hline
\multicolumn{1}{l|}{$\text{BERT}_{\text{base}}$ [CLS]$^\dagger$}  &   20.16  &   30.01 &  20.09   &  36.88   &  38.08  & 16.05  & \multicolumn{1}{c|}{42.63}  &  29.19 \\ \hline
\multicolumn{1}{l|}{$\text{BERT}_{\text{base}}$-flow}  &   59.54 & 64.69 &  64.66  &  72.92  &  71.84 &  58.56  & \multicolumn{1}{c|}{65.44}  &  65.38 \\ \hline
\multicolumn{1}{l|}{IS-$\text{BERT}_{\text{base}}$}    &  56.77   &   69.24  &    61.21  &  75.23  &   70.16 &   69.21    & \multicolumn{1}{c|}{64.25} &  66.58 \\ \hline
\multicolumn{1}{l|}{$\text{BERT}_{\text{base}}$-whitening$^\spadesuit$}   & 61.46 &  66.71  &  66.17  &  74.82 &   72.10  &  67.51  & \multicolumn{1}{c|}{64.90}  & 67.67 \\ \hline \hline
\multicolumn{1}{l|}{MoCo-$\text{BERT}_{\text{base}}$} & \textbf{68.85}  & \textbf{77.52} &  \textbf{75.85}  &  \textbf{83.14} & \textbf{80.15}  &  \textbf{77.50}  & \multicolumn{1}{c|}{\textbf{72.48}}  & \textbf{76.50} \\ \hline
\multicolumn{9}{l}{\textit{w/ labeled NLI supervision}}   \\ \hline
\multicolumn{1}{l|}{InferSent}      &  52.86   &  66.75 &   62.15   &  72.77  &   66.87 &  68.03  & \multicolumn{1}{c|}{65.65}  &  65.01 \\ \hline
\multicolumn{1}{l|}{S$\text{BERT}_{\text{base}}$-NLI$^\dagger$}   & 68.70 & 74.37  &  74.73  & 79.65  & 75.21 &  77.63  & \multicolumn{1}{c|}{74.84}  & 75.02 \\ \hline
\multicolumn{1}{l|}{$\text{BERT}_{\text{base}}$-flow}      &   67.75  &  76.73  &  75.53 &  80.63 &  77.58 &  79.10  & \multicolumn{1}{c|}{78.03}  &  76.48 \\ \hline
\multicolumn{1}{l|}{$\text{BERT}_{\text{base}}$-whitening$^\spadesuit$}   & 69.87 &  77.11  &  76.13   &  82.73 &   78.08  &  79.16  & \multicolumn{1}{c|}{76.44}  & 77.07 \\ \hline \hline
\multicolumn{1}{l|}{MoCo-$\text{BERT}_{\text{base}}$+NLI}   &  \textbf{71.66}  & \textbf{79.42} & \textbf{76.37}  &  \textbf{84.08}  &  \textbf{80.81} & \textbf{82.15}  & \multicolumn{1}{c|}{\textbf{78.19}}   &  \textbf{78.95} \\ \hline
\end{tabular}}
\caption{Spearman's correlation for 7 STS datasets downloaded from SentEval ~\citep{Conneau2018SentEvalAE}.
We report ``weighted mean'' (wmean) from SentEval toolkit.
Baseline systems include $\text{BERT}_{\text{base}}$-flow ~\citep{li2020sentence},
IS-$\text{BERT}_{\text{base}}$ ~\citep{zhang2020unsupervised},
$\text{BERT}_{\text{base}}$-whitening$^\spadesuit$ ~\citep{Su2021WhiteningSR},
and InferSent ~\citep{conneau2017supervised}.
$\dagger$: from ~\citet{Reimers2019SentenceBERTSE}.}
\label{tab:sts}
\end{table*}

We evaluate the performance of MoCo-BERT for STS without training on any labeled STS data,
following the procedure by ~\citet{Reimers2019SentenceBERTSE}.
All results are based on $\text{BERT}_{\text{base}}$.
Given a pair of English sentences,
the semantic similarity is computed with a simple dot product.
We also report the results using labeled \emph{natural language inference} (NLI) data.
A two-layer MLP with $256$ hidden units and a 3-way classification head
is added on top of the sentence representations.
The training set of SNLI ~\citep{bowman2015large} and MNLI ~\citep{williams2018broad}
are used for multi-task training.
See Appendix ~\ref{app:mt_nli} for the detailed setup.

As pointed out by ~\citet{gao2021simcse},
existing works follow inconsistent evaluation protocols,
and thus may cause unfair comparison.
We report results for both ``weighted mean'' (wmean) and ``all'' settings ~\citep{gao2021simcse}
in Table ~\ref{tab:sts} and ~\ref{tab:all_sts} respectively.

When training on translation ranking task only,
MoCo-BERT improves the average correlation from $67.67$ to $76.50$ ($+8.83$).
With labeled NLI supervision,
MoCo-BERT+NLI advances state-of-the-art from $77.07$ to $78.95$ ($+1.88$).

\subsection{Model Analysis}
We conduct a series of experiments to better understand
the behavior of MoCo-BERT.
Unless explicitly mentioned,
we use a memory queue size $204800$ for efficiency.
\newline

\begin{figure}[ht]
\begin{center}
 \includegraphics[width=0.99\linewidth]{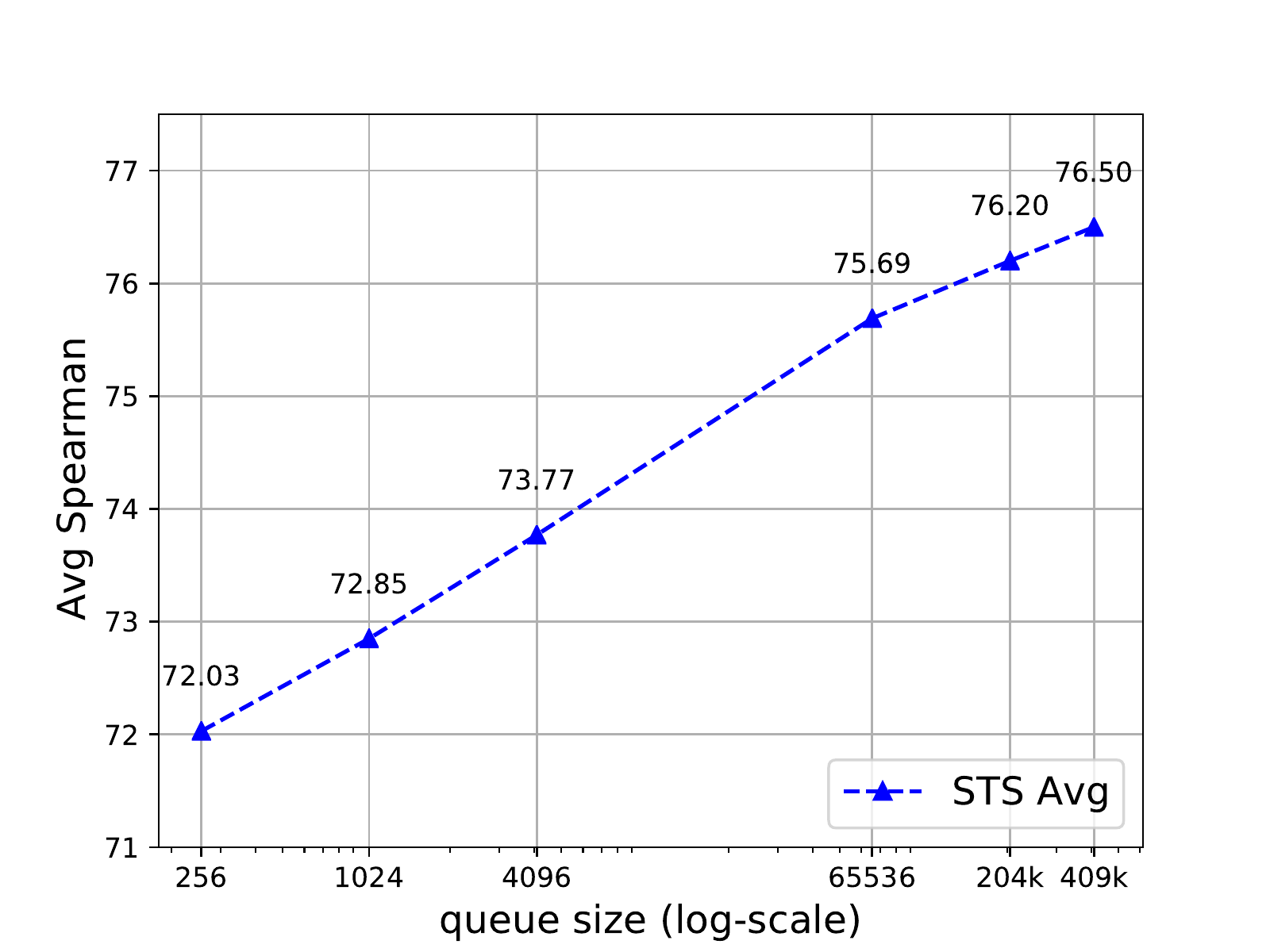}
 \caption{Average Spearman's correlation across 7 STS datasets for different memory queue sizes.
 The performance does not seem to saturate with queue size as large as $409k$.
 We do not run experiments $> 409k$ as it reaches the GPU memory limit.}
 \label{fig:queue}
\end{center}
\end{figure}

\noindent
\textbf{Memory queue size }
One primary motivation of MoCo is to introduce more negatives
to improve the quality of the learned representations.
In Figure ~\ref{fig:queue},
as expected,
the performance consistently increases as the memory queue becomes larger.
For visual representation learning,
the performance usually saturates with queue size $\sim 65536$ ~\citep{he2020momentum},
but the ceiling is much higher in our case.
Also notice that the model can still reach $72.03$ with a small batch size $256$,
which might be because the encoders have already been pre-trained with MLM.
\newline

\begin{table}[ht]
\centering
\begin{tabular}{ccccc}
\hline
Temperature & 0.01 & 0.04 & 0.07 & 0.1 \\ \hline
STS Avg   & 74.80 & \textbf{76.20}   & 74.23  & 69.81  \\ \hline
BUCC F1   & 90.76 & \textbf{93.14}   & 90.42  & 77.04  \\ \hline
\end{tabular}
\caption{Performance of our proposed MoCo-BERT under different temperatures.}
\label{tab:temperature}
\end{table}

\noindent
\textbf{Temperature }
A lower temperature $\tau$ in InfoNCE loss makes the model
focus more on the hard negative examples,
but it also risks over-fitting label noises.
Table ~\ref{tab:temperature} shows that $\tau$
could dramatically affect downstream performance,
with $\tau=0.04$ getting the best results on both STS and BUCC bitext mining tasks.
The optimal $\tau$ is likely to be task-specific.

\begin{table}[ht]
\centering
\begin{tabular}{l|cc}
\hline
Model        & STS Avg & BUCC F1 \\ \hline
MoCo-BERT    & \textbf{76.20} & \textbf{93.14} \\ \hline
w/o momentum & -0.01 &  0.00 \\ \hline
\end{tabular}
\caption{Ablation results for momentum update mechanism.
\emph{w/o momentum} shares the parameters between the momentum encoder and the base encoder.}
\label{tab:momentum}
\end{table}

\noindent
\textbf{Momentum Update }
We also empirically verify if the momentum update mechanism is really necessary.
Momentum update provides a more consistent matching target
but also complicates the training procedure.
In Table ~\ref{tab:momentum},
without momentum update,
the model simply fails to converge with the training loss oscillating back and forth.
The resulting Spearman's correlation is virtually the same as random predictions.
\newline

\begin{table}[ht]
\centering
\begin{tabular}{l|cc}
\hline
Pooling          & STS Avg & BUCC F1 \\ \hline
mean pooling & \textbf{76.20} & \textbf{93.14} \\ \hline
max pooling  & 75.90  & 92.78 \\ \hline
{[}CLS{]}   & 75.97 & 92.47 \\ \hline
\end{tabular}
\caption{Performance comparison between different pooling mechanisms for MoCo-BERT.}
\label{tab:pooling}
\end{table}

\noindent
\textbf{Pooling mechanism }
Though the standard practices of fine-tuning BERT ~\citep{devlin-etal-2019-bert}
directly use hidden states from [CLS] token,
~\citet{Reimers2019SentenceBERTSE,li2020sentence} have shown
that pooling mechanisms matter for downstream STS tasks.
We experiment with mean pooling, max pooling, and [CLS] embedding,
with results listed in Table ~\ref{tab:pooling}.
Consistent with ~\citet{Reimers2019SentenceBERTSE},
mean pooling has a slight but pretty much negligible advantage
over other methods.

In Appendix ~\ref{app:visualize},
we also showcase some visualization and sentence retrieval results.

\section{Related Work}

\noindent
\textbf{Multilingual representation learning }
aims to jointly model multiple languages.
Such representations are crucial for multilingual neural machine translation ~\citep{aharoni2019massively},
zero-shot cross-lingual transfer ~\citep{Artetxe2019MassivelyMS},
and cross-lingual semantic retrieval ~\citep{Yang2020MultilingualUS} etc.
Multilingual BERT ~\citep{Pires2019HowMI} simply pre-trains
on the concatenation of monolingual corpora
and shows good generalization for tasks
like cross-lingual text classification ~\citep{Hu2020XTREMEAM}.
Another line of work explicitly aligns representations from language-specific models,
either unsupervised ~\citep{lample2018word} or supervised ~\citep{reimers2020making,feng2020language}.
\newline

\noindent
\textbf{Contrastive learning }
works by pulling positive instances closer and pushing negatives far apart.
It has achieved great successes in self-supervised vision representation learning,
including SimCLR ~\citep{chen2020simple},
MoCo ~\citep{he2020momentum,chen2020improved},
BYOL ~\citep{grill2020bootstrap},
CLIP ~\citep{radford2021learning} etc.
Recent efforts introduced contrastive learning into various NLP tasks
~\citep{xiong2020approximate,giorgi2020declutr,chi2021infoxlm,gunel2020supervised}.
Concurrent to our work,
SimCSE ~\citep{gao2021simcse} uses dropout and hard negatives from NLI datasets
for contrastive sentence similarity learning,
Sentence-T5 ~\citep{ni2021sentence} outperforms SimCSE by scaling to larger models,
and xMoCo ~\citep{yangxmoco} adopts a similar variant of MoCo for open-domain question answering.
\newline

\noindent
\textbf{Semantic textual similarity }
is a long-standing NLP task.
Early approaches ~\citep{seco2004intrinsic,budanitsky2001semantic} use lexical resources such as WordNet
to measure the similarity of texts.
A series of SemEval shared tasks ~\citep{agirre2012semeval,agirre2014semeval}
provide a suite of benchmark datasets that is now widely used for evaluation.
Since obtaining large amounts of high-quality STS training data is non-trivial,
most STS models are based on weak supervision data,
including conversations ~\citep{yang2018learning},
NLI ~\citep{conneau2017supervised,Reimers2019SentenceBERTSE},
and QA pairs ~\citep{ni2021sentence}.

\section{Conclusion}
This paper proposes a novel method that aims to solve the \emph{easy negatives}
issue to better align cross-lingual sentence representations.
Extensive experiments on multiple cross-lingual and monolingual evaluation datasets
show the superiority of the resulting representations.
For future work,
we would like to explore other contrastive learning methods ~\citep{grill2020bootstrap,xiong2020approximate},
and experiment with more downstream tasks
including paraphrase mining,
text clustering,
and bilingual lexicon induction etc.

\section*{Acknowledgements}
We would like to thank three anonymous reviewers for their valuable comments,
and EMNLP 2021 organizers for their efforts.
We also want to thank Yueya He for useful suggestions on an early draft of this paper.

% Entries for the entire Anthology, followed by custom entries
\bibliography{anthology,custom}
\bibliographystyle{acl_natbib}

\appendix

\section{Details on Training Data and Hyperparameters} ~\label{app:setup}

\begin{table}[ht]
\centering
\scalebox{0.9}{\begin{tabular}{l|cc}
\hline
Dataset & \# of sents & \# of sampled \\ \hline
Tatoeba   &   46k &  46k  \\ \hline
News Commentary  &   320k &   320k  \\ \hline
UNCorpus        &   16M  &   1M  \\ \hline
CWMT-neu2017    &   2M   &  2M  \\ \hline
CWMT-casia2015    &  1M  &   1M \\ \hline
CWMT-casict2015   &  2M   &   1M  \\ \hline
\end{tabular}}
\caption{List of parallel corpora used.
\emph{\# of sampled} are randomly sampled subset from the corresponding dataset
to make the training cost manageable.
Duplicates are removed during preprocess.}
\label{tab:train_corpora}
\end{table}

\begin{table*}[ht]
\centering
\scalebox{0.98}{\begin{tabular}{ccccccccc}
\hline
\multicolumn{1}{l|}{Model} & STS-12 & STS-13 & STS-14 & STS-15 & STS-16 & STS-B & \multicolumn{1}{c|}{SICK-R} & Avg \\ \hline
\multicolumn{9}{l}{\textit{w/o labeled NLI supervision}}         \\ \hline
\multicolumn{1}{l|}{$\text{BERT}_{\text{base}}$-flow}  &   58.40 & 67.10 &  60.85  & 75.16  &  71.22  & 68.66  & \multicolumn{1}{c|}{64.47}  &  66.55 \\ \hline
\multicolumn{1}{l|}{$\text{BERT}_{\text{base}}$-whitening}   & 57.83  & 66.90  &  60.90  &  75.08 &   71.31  &  68.24  & \multicolumn{1}{c|}{63.73}  & 66.28 \\ \hline \hline
\multicolumn{1}{l|}{MoCo-$\text{BERT}_{\text{base}}$} & \textbf{70.99}  & \textbf{76.51} &  \textbf{73.17}  &  \textbf{82.09} & \textbf{78.32}  &  \textbf{77.50}  & \multicolumn{1}{c|}{\textbf{72.48}}  & \textbf{75.87} \\ \hline
\multicolumn{9}{l}{\textit{w/ labeled NLI supervision}}   \\ \hline
\multicolumn{1}{l|}{S$\text{BERT}_{\text{base}}$-NLI}   & 70.97 &  76.53  &  73.19   &  79.09 &   74.30  &  77.03  & \multicolumn{1}{c|}{72.91}  & 74.89 \\ \hline
\multicolumn{1}{l|}{$\text{BERT}_{\text{base}}$-flow}   &   69.78  & 77.27  &  74.35  & 82.01  &  77.46  & 79.12  & \multicolumn{1}{c|}{76.21}  &  76.60 \\ \hline
\multicolumn{1}{l|}{$\text{BERT}_{\text{base}}$-whitening}   & 69.65  &  77.57  &  \textbf{74.66}  & 82.27 &   78.39  & 79.52  & \multicolumn{1}{c|}{76.91}  & 77.00 \\ \hline \hline
\multicolumn{1}{l|}{MoCo-$\text{BERT}_{\text{base}}$+NLI}   &  \textbf{76.07}  & \textbf{78.33} & 74.51  &  \textbf{84.19}  &  \textbf{78.74} & \textbf{82.15}  & \multicolumn{1}{c|}{\textbf{78.19}}   &  \textbf{78.88} \\ \hline
\end{tabular}}
\caption{Spearman's correlation for 7 STS datasets under the ``all'' evaluation setting ~\citep{gao2021simcse}.
We use the official script from SimCSE.}
\label{tab:all_sts}
\end{table*}

\begin{table}[ht]
\centering
\scalebox{0.9}{\begin{tabular}{l|c}
\hline
Hyperparameter & value \\ \hline
\# of epochs  &  15 \\ \hline
\# of GPUs  &  4 \\ \hline
queue size & 409k  \\ \hline
temperature $\tau$  &   0.04 \\ \hline
momentum coefficient  &  0.999  \\ \hline
learning rate  &  $4\times10^{-5}$ \\ \hline
gradient clip & 10  \\ \hline
warmup steps & 400  \\ \hline
batch size  & 1024 \\ \hline
dropout & 0.1 \\ \hline
weight decay  & $10^{-4}$ \\ \hline
pooling  &  mean \\ \hline
\end{tabular}}
\caption{Hyperparameters for our proposed model.}
\label{tab:appendix_hyper}
\end{table}

We list all the parallel corpora used by this paper in Table ~\ref{tab:train_corpora}.
Hyperparameters are available in Table ~\ref{tab:appendix_hyper}.
We start with the default hyperparameters from MoCo ~\citep{he2020momentum}
and use grid search to find the optimal values for several hyperparameters.
The specific search ranges are \{$10^{-5}$, $2\times10^{-5}$, $4\times10^{-5}$\} for learning rate,
\{$102k$, $204k$, $409k$\} for queue size,
\{$0.01$, $0.04$, $0.07$, $0.1$\} for temperature,
and \{$0.9999$, $0.999$, $0.99$\} for momentum coefficient.
The entire training process takes approximately 15 hours
with 4 V100 GPUs and automatic mixed precision support from PyTorch.

\section{Multi-task with NLI} ~\label{app:mt_nli}
Given a premise $x_p$ and a hypothesis $x_h$,
the sentence representations are computed as stated in the paper.
Then,
a two-layer MLP with $256$ hidden units, ReLU activation,
and a 3-way classification head
is added on top of the sentence representations.
Dropout $0.1$ is applied to the hidden units.
The loss function $\text{L}_{\text{nli}}(x_p, x_h)$
is simply the cross-entropy between gold label and softmax outputs.
The model is jointly optimized with the following:

\begin{equation}
    \min\ \ \text{L}(x, y) + \text{L}(y, x) + \alpha \text{L}_{\text{nli}}(x_p, x_h)
\end{equation}

Where $\alpha$ is used to balance different training objectives,
we set $\alpha=0.1$ empirically.
The batch size for NLI loss is $128$.
The training set is the union of SNLI ~\citep{bowman2015large}
and MNLI ~\citep{williams2018broad} dataset (\textasciitilde1M sentence pairs).

\section{Visualization of Sentence Representations} ~\label{app:visualize}

\begin{figure*}[ht]
\begin{center}
 \includegraphics[width=1.0\linewidth]{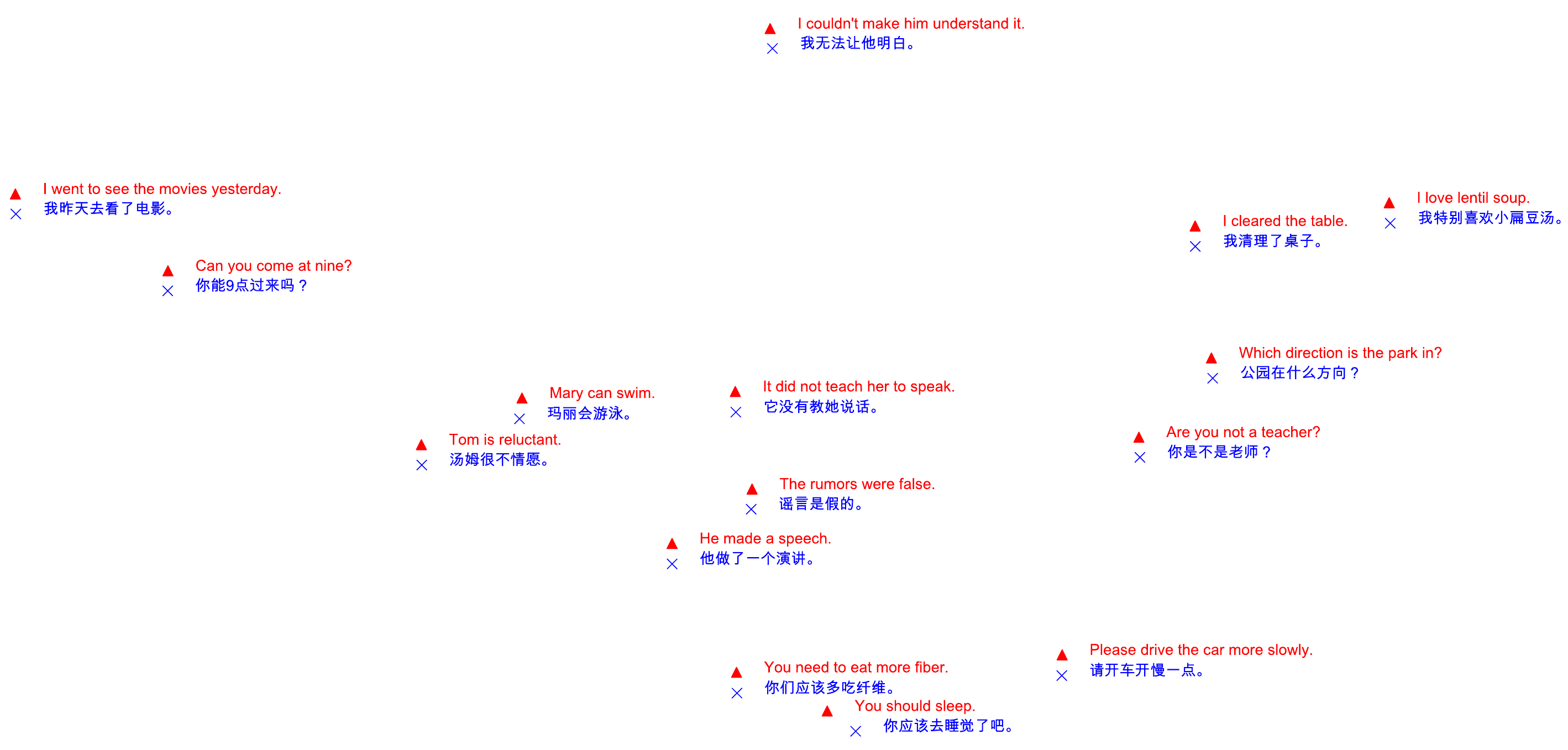}
 \caption{t-SNE visualization of the representations of 15 random parallel sentences from Tatoeba test set.
 For visualization purpose,
 if two points are too close,
 we move them a little bit far apart.
 Enlarge the graph for better views.}
 \label{fig:t_sne}
\end{center}
\end{figure*}

\begin{table*}[ht!]
\centering
\begin{tabular}{l|l}
\hline
\multicolumn{2}{l}{\textbf{query}: \emph{I am willing to devote my life to education career.}} \\ \hline \hline
\multicolumn{1}{l|}{0.853} & He dedicated his life to the cause of education. \\ \hline
\multicolumn{1}{l|}{0.776}    & He devoted his whole life to education. \\ \hline
\multicolumn{1}{l|}{0.764}    &  She has dedicated herself to the cause of education.  \\ \hline \hline
\multicolumn{2}{l}{\textbf{query}: \emph{The Committee resumed consideration of the item.}}                    \\ \hline \hline
\multicolumn{1}{l|}{0.928} & The Committee continued consideration of the item.   \\ \hline
\multicolumn{1}{l|}{0.843} & The Committee resumed its consideration of this agenda item. \\ \hline
\multicolumn{1}{l|}{0.686} & The Committee began its consideration of the item. \\ \hline \hline
\multicolumn{2}{l}{\textbf{query}: \emph{There are a great many books on the bookshelf.}} \\ \hline \hline
\multicolumn{1}{l|}{0.837} & There are many books on the bookcase.   \\ \hline
\multicolumn{1}{l|}{0.690} & There is a heap of books on the table. \\ \hline
\multicolumn{1}{l|}{0.655} & The bookshelf is crowded with books on different subjects. \\ \hline \hline
\multicolumn{2}{l}{\textbf{query}: \emph{Everyone has the privilege to be tried by a jury.}} \\ \hline \hline
\multicolumn{1}{l|}{0.718} & They have the right to have their case heard by a jury. \\ \hline
\multicolumn{1}{l|}{0.647} & Every defendant charged with a felony has a right to be charged by the Grand Jury. \\ \hline
\multicolumn{1}{l|}{0.580} & Everyone has the right to be educated. \\ \hline
\end{tabular}
\caption{Examples of sentence retrieval using learned representations.
Given a query,
we use cosine similarity to retrieve the $3$ nearest neighbors (excluding exact match).
The first column is the cosine similarity score between the query and retrieved sentences.
The corpus is 1M random English sentences from the training data.}
\label{tab:sent_retrieval}
\end{table*}

To visualize the learned sentence representations,
we use t-SNE ~\citep{Maaten2008VisualizingDU} for dimensionality reduction.
In Figure ~\ref{fig:t_sne},
we can see the representations of parallel sentences are very close,
indicating that our proposed model is successful at aligning cross-lingual representations.

In Table ~\ref{tab:sent_retrieval},
we illustrate the results of monolingual sentence retrieval.
Most top-ranked sentences indeed share similar semantics with the given query,
this paves the way for potential applications like paraphrase mining.

\end{document}